\pdfoutput=1
\documentclass[11pt]{article}

\usepackage[]{emnlp}

\usepackage{times}
\usepackage{latexsym}
\usepackage{tabularx}
\usepackage{graphicx}
\usepackage{amsmath,amssymb}
\usepackage{subfig}
\usepackage{stmaryrd}
\usepackage{multirow}
\usepackage{titlesec}

\newcommand\sA{\mathcal{A}}
\newcommand\sP{\mathcal{P}}
\newcommand\sE{\mathcal{E}}
\newcommand\sR{\mathcal{R}}
\newcommand\sK{\mathcal{K}}
\newcommand\den[1]{\llbracket #1 \rrbracket}

\newcommand\commentout[1]{}

\usepackage[T1]{fontenc}
\usepackage{xcolor}
\usepackage{booktabs}
\usepackage[utf8]{inputenc}

\usepackage{microtype}

\newcommand\ourbenchmark{\textsc{QAMParI}}
\newif\ifcomments
\commentstrue
\ifcomments
    \providecommand{\jb}[1]{{\protect\color{blue}{[JB: #1]}}}
    \providecommand{\sa}[1]{{\protect\color{red}{[SA: #1]}}} 
    \providecommand{\tw}[1]{{\protect\color{magenta}{[TW: #1]}}} 
    \providecommand{\oy}[1]{{\protect\color{orange}{[OY: #1]}}} 
     
    \providecommand{\oh}[1]{{\protect\color{cyan}{[OR: #1]}}} 

\else
    \providecommand{\jb}[1]{}
    \providecommand{\sa}[1]{}    
    \providecommand{\tw}[1]{}    
    \providecommand{\oy}[1]{}    
    \providecommand{\oh}[1]{}    
\fi

\newif\ifcomments
\commentstrue
\ifcomments
    
    \def \ifempty#1{\def\temp{#1} \ifx\temp\empty }

\title{\ourbenchmark{}: A Benchmark for Open-domain Questions with Many Answers}

\author{Samuel Joseph Amouyal \hspace{0.5cm} Tomer Wolfson \hspace{0.5cm} Ohad Rubin \hspace{0.5cm} Ori Yoran  \\ \bf{Jonathan Herzig} \hspace{0.5cm} \bf{Jonathan Berant} \\
  Blavatnik School of Computer Science, Tel Aviv University, Israel \\
 \texttt{\{samuel.amouyal, ohad.rubin, joberant\}@cs.tau.ac.il}}

\begin{document}
\maketitle

\begin{abstract}
% TW: wrote a slightly revised the abstract below. Feel free to change/discard/revise as you see fit:
Existing benchmarks for open-domain question answering (ODQA) typically focus on questions whose answers are all in a single paragraph. By contrast, many natural questions, such as \emph{``What players were drafted by the Brooklyn Nets?''} have a long \emph{list of answers} extracted from multiple paragraphs. Answering such questions requires retrieving and reading 
many passages from a large corpus. 
We introduce \ourbenchmark{}, an ODQA benchmark, where answers are lists of entities, spread across many paragraphs. 
We created \ourbenchmark{} by (a) generating questions with multiple answers from Wikipedia's knowledge graph and tables, (b) automatically pairing answers with supporting evidence in Wikipedia paragraphs, and (c) manually paraphrasing questions and validating each answer. Across a wide range of ODQA models, we find that \ourbenchmark{} is challenging in terms of both passage retrieval and answer generation, with models reaching an F$_1$ score of 32.8 at best. 
We view \ourbenchmark{} as a valuable resource for ODQA research, which will aid to develop models that handle a broad range of question types, including single and multi-answer questions.

\end{abstract}

\section{Introduction}
\label{sec:intro}

Open-domain question answering (ODQA) is a core language understanding task concerned with answering factoid questions over large document collections \cite{trec_qa,brill-etal-2002-analysis}. Due to its wide applicability, ODQA has received substantial attention in recent years 
\cite{chen_2017_odqa,lee-etal-2019-latent,dpr_article}. Typically, systems tackling ODQA tasks follow the ``retrieve-and-read'' paradigm, where \emph{a retriever} first retrieves a set of candidate passages, followed by \emph{a reader} which receives the retrieved passages and produces the final answer.

\begin{figure}[t]
    \centering
    \includegraphics[width=7.7cm,height=5.5cm]{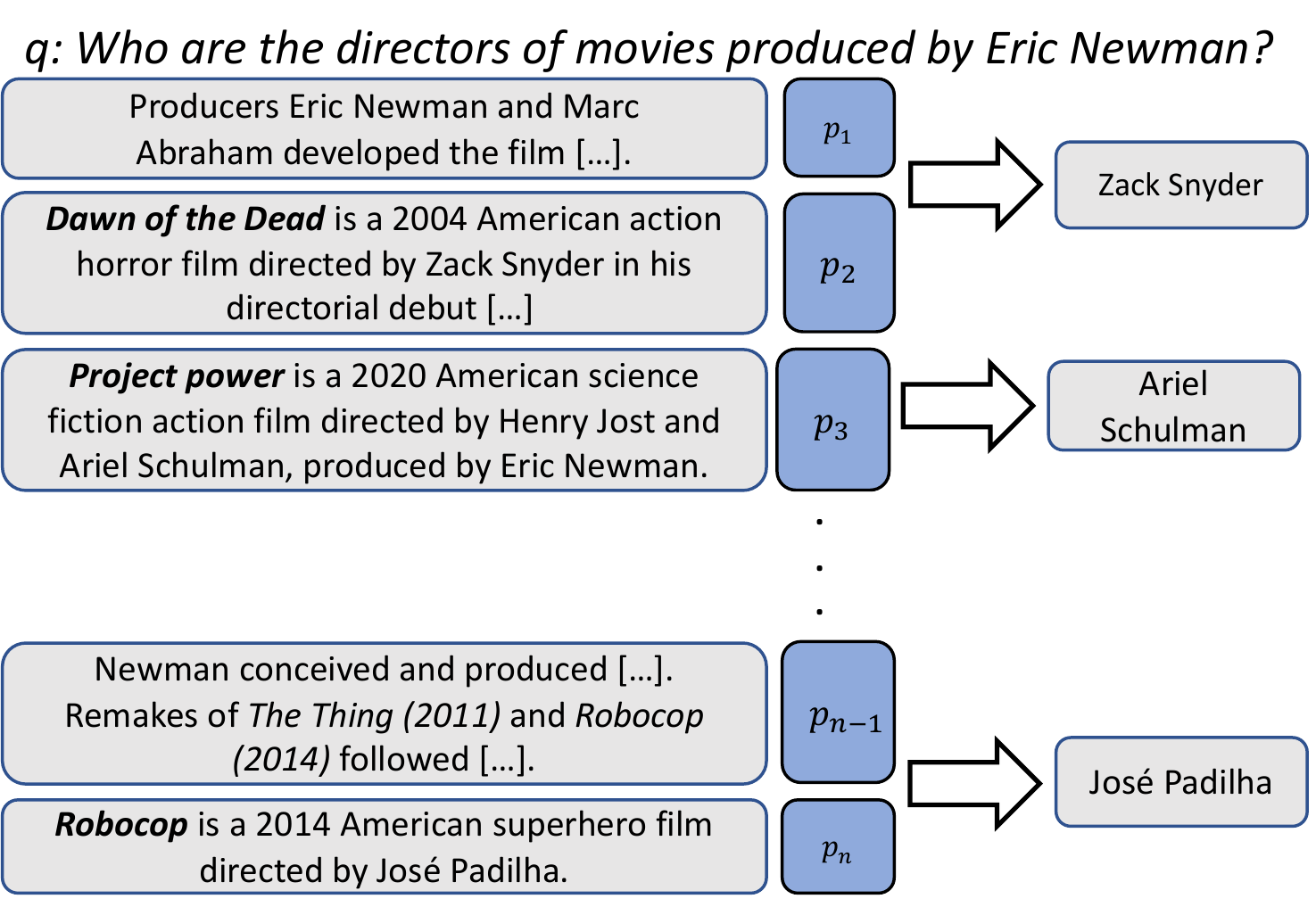}
    \caption{An example from \ourbenchmark{} with a generated question $q$, a subset of its evidence Wikipedia passages (left, $p_i$) and their corresponding answer.}
    \label{fig:figure1}
\end{figure}

The retrieve-and-read paradigm has been effective for benchmarks such as Natural Questions (NQ) \cite{natural_question_qa} and TriviaQA \cite{trivia_qa}, where the answer is a single phrase from a single passage. However, in many cases, a question might have \emph{many} answers, spread across multiple passages.
Consider the example in Fig.~\ref{fig:figure1}. Eric Newman produced multiple movies, so finding them, along with their directors, requires incorporating information from many passages. 
Such questions pose two main challenges to retrieve-and-read systems. First, as there are multiple answers that can be far apart, the reader must reason over a long text sequence to generate all correct answers. Second, since the reader is computationally constrained to process at most K passages, the retriever must score all necessary passages at its top-K results, which is challenging and even impossible when the number of relevant passages exceeds K.

Nevertheless, research on multi-answer questions has largely been underexplored. While previous works proposed questions that involve reading multiple passages, the number of passages was quite small. \textsc{AmbigQA} \cite{ambig_qa} studied ambiguous questions from NQ with several answers. However, 
as 70\% of its questions have at most two answers, retrieve-and-read models can be adapted to \textsc{AmbigQA}.
\textsc{HotpotQA} \cite{hotpotqa} focused on multi-hop reasoning, but its questions require no more than two passages to answer.
\textsc{WikiNLDB} \cite{nrot_qa} is a benchmark for testing reasoning over multiple facts. However, \textsc{WikiNLDB} restricted its text corpus to databases of 1,000 facts at most, making it significantly smaller than standard ODQA corpora. Moreover, these facts are model-generated utterances rather than natural language passages. 
% \edit{}{However, there are many interesting questions with many answers (\emph{Which drugs are effective against skin cancer?}, \emph{Which plants can be grown in apartment?}). Google itself increasingly supports retrieving HTML tables and lists, when posed with multi-answer questions. We believe that this type of question does not appear often in datasets with questions from real-world user queries \cite{ms_marco, natural_question_qa} due to some performance bias, with people asking questions they know they can get the answer to. Answering questions with many answers is an auxiliary task that models should also be able to perform.}
Multi-answer questions are also rare in real-world user questions \cite{ms_marco, natural_question_qa}, which can be attributed to the performance bias of existing systems. Namely, people mostly pose questions that they can successfully get answers to with current technology. This does not diminish the importance of multi-answer questions (\emph{`Which drugs are effective against skin cancer?'}; 
\emph{`Which plants can be grown in an apartment?'}), which constitute an important research challenge.

%\tw{slightly rephrased: Unfortunately, multi-answer questions do not often appear in real-world user questions \cite{ms_marco, natural_question_qa}. We attribute this to the performance bias of existing systems. Namely, people naturally pose questions that they can successfully get answers to. This does not diminish the potential benefit of multi-answer questions (\emph{Which drugs are effective against skin cancer?}; 
%\emph{Which plants can be grown in apartment?}). We therefore view multi-answer QA as an important task for existing models.}

In this work, we present \ourbenchmark{}, a benchmark for \textbf{Q}uestions with many \textbf{A}nswers over \textbf{M}ultiple \textbf{Par}agraphs, \textbf{I}ndeed. All questions in \ourbenchmark{} have at least 5 answers, with an average of 13 answers. Examples are semi-automatically generated using two data sources, Wikidata \cite{vrandevcic2014Wikidata} and Wikipedia tables.
We automatically generate multi-answer questions of the form \emph{``What/Who has [relation] with [entity]?''} and convert these into pseudo-language using manually defined templates. Then, we verify that our questions are answerable from Wikipedia by automatically extracting evidence passages for all their answers. Finally, we use crowdsourcing to validate example correctness, and paraphrase questions into natural language \cite{wang2015overnight}. To further enrich our data %increase the richness of questions,
we also generate \emph{composition} questions, that compose two relations (as in Fig.~\ref{fig:figure1}), and \emph{intersection} questions, such as \emph{``What movies were produced and directed by Clint Eastwood?''}. Overall, \ourbenchmark{} contains 2K development and test questions and more than 60K training examples -- see Tab.~\ref{tab:examples} for some examples. %\tw{We describe our question generation process, and provide examples, in Section 2}

We evaluate a large suite of baselines, including models from the retrieve-and-read family as well as  a closed-book question answering model \cite{roberts-etal-2020-much},
and find that they struggle on \ourbenchmark{}. In the retrieve-and-read setup,
we experiment with both BM25 \cite{robertson2009bm25} and DPR \cite{dpr_article} retrievers, followed by
either (a) a RAG-like reader \cite{RAG} that given each retrieved passage either decodes an answer or abstains, or (b) an
FiD reader \cite{fid_article} that takes the encoded representations of multiple passages and decodes the list of answers directly. 

%FiD \cite{fid_article} \edit{}{and a RAG alike model \cite{RAG}}, \edit{a}{} state-of-the-art ODQA model\edit{}{s} from the retrieve-and-read family, and find that \edit{it}{they} struggle\edit{s}{} on \ourbenchmark{}. Specifically, we use a BM25 \cite{robertson2009bm25} and a DPR \cite{dpr_article} retriever followed by an FiD reader \cite{fid_article} that, given encoded representations of many passages, directly decodes the answer list \edit{}{or a RAG alike reader that decodes from each passage an answer}. To evaluate our models we compare their predicted set of answers to the gold set of answers.

When training models on \ourbenchmark{} alone, or in a multi-task setup with NQ, we observe that \ourbenchmark{} is challenging in terms of both passage retrieval and answer generation. Namely, the best model reaches an F$_1$ score of 32.8. Moreover, models return more than 80\% of the correct answers in only 31.2\% of the test examples, well below performance on single-answer datasets like NQ.

To summarize, \ourbenchmark{} is a challenging benchmark for evaluating the ability of ODQA models to handle questions with many answers over multiple passages.
We advocate to evaluate ODQA models not on \ourbenchmark{} alone, but alongside benchmarks such as NQ and TriviaQA. Such joint evaluation will test models' ability to handle both single- and multi-answer questions, an evaluation that the community is currently lacking.
The \ourbenchmark{} benchmark, models and relevant codebase are available at: \url{https://samsam3232.github.io/qampari/}. 
\section{Dataset Construction}
\label{sec:dataset_concstrution}

Each example in \ourbenchmark{} is a triple $(q, \sA, \sP)$, where $q$ is a question, $\mathcal{A}$ is a set of answers and $\mathcal{P}$ is a set of passages from our target corpus. An answer $a \in \sA$ has 1-2 evidence passages from $\sP$ (see Fig.~\ref{fig:figure1}). 

We define passages as consecutive sentences from our corpus (Wikipedia), that span on average 100 words. As our focus is multi-answer questions, examples in \ourbenchmark{} have $|\sA| \geq 5$.

\paragraph{Overview}
We generate examples in two steps. First, we generate \emph{simple questions} that involve a single entity and relation, e.g., \emph{``Who was drafted by the Brooklyn Nets?''} (\S\ref{subsec:simple_quest}). Then, we expand such questions to generate \emph{complex questions} with \emph{intersection} and \emph{composition} operations (\S\ref{subsec:complex_questions}).

To increase diversity, questions are generated from two data sources, Wikidata and Wikipedia tables. We first describe example generation over Wikidata, then briefly present the generation process from Wikipedia tables in \S\ref{subsec:wikipedia_tables}. 
In both cases, we ensure answers can be derived from evidence passages in Wikipedia.\footnote{Wikipedia dump: 2021-08-01}
Tab.~\ref{tab:examples} presents examples from each data source and question type.

\begin{table*}[t]
  \centering
  \footnotesize
  \renewcommand{\arraystretch}{1.1}
  \begin{tabular}{c l p{7.5cm} l}
  \toprule
    \textbf{Data source} & \textbf{Question type} & \textbf{Question} & \textbf{Answer example} \\
    \midrule
    \multirow{3}{*}{\textbf{Wikidata}} & Simple & Who is or was a member of the Australian Army? & George Macarthur-Onslow \\
    & Intersection & What movie produced by Jerry Ward was also directed by Vincent Sherman? & Hard Way \\
    & Composition & From which country did Seattle Storm make draft selections? & Australia \\ 
    \hline
    \multirow{2}{*}{\textbf{Wiki. tables}} & Simple & What magazine is a satirical magazine? & The Clinic \\
    & Composition & What are the museums found in Concord, Massachusetts? & The Wayside \\
 \bottomrule
  \end{tabular}
  \caption{Example questions and one representative answer for all data sources and question types.}
  \label{tab:examples}
\end{table*}

\begin{figure*}[t]
    \centering
    \includegraphics[width=15.3cm, height=3cm]{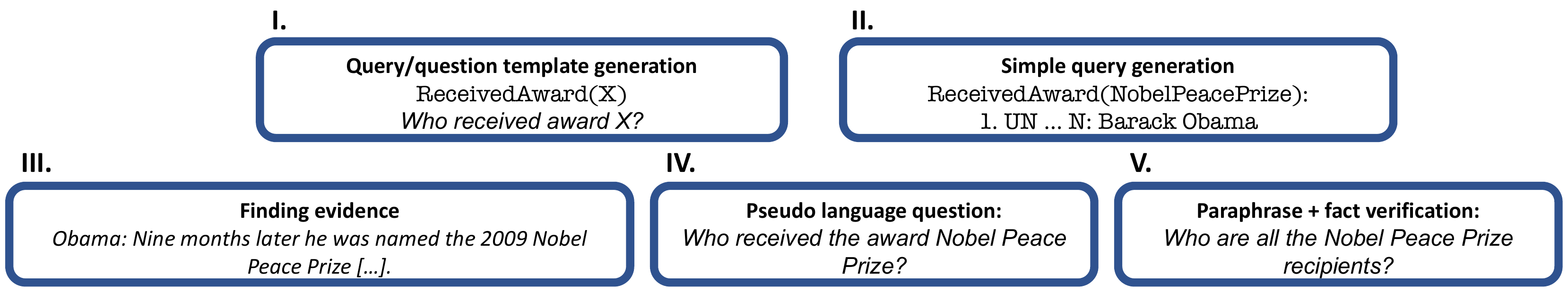}
    \caption{An overview of example generation for simple questions.}
    \label{fig:figure2}
\end{figure*}

\paragraph{Notation}
We introduce notation for formal queries over Wikidata to explain example generation.
Wikidata is a knowledge graph, $\sK$, that can be viewed as a set of labeled edges $(e_1, r, e_2)$. Graph nodes $e_1, e_2 \in \sE$ are entities connected by an edge labeled by the relation $r \in \sR$. For example, a possible labeled edge is (\texttt{BarackObama}, \texttt{ReceivedAward}, \texttt{NobelPeacePrize}). 

One can query $\sK$ by applying a relation $r$ over an entity $e$, resulting in a \emph{simple query} $r(e)$ whose \emph{denotation} (answer set) is $\den{r(e)} = \{e_i \mid (e_i, r, e) \in \sK\}$.
\emph{Composition queries} are formed by applying a relation over the result of a simple query. We denote a composition query by $r_2(r_1(e))$, and its denotation is $\den{r_2(r_1(e))} = \{e_i \mid \exists e_j  \textit{ s.t }(e_i, r_2, e_j) \in \sK\ \land (e_j, r_1, e) \in \sK \}$. 
Last, an \emph{intersection query} $r_1(e_1) \sqcap r_2(e_2)$ corresponds to the intersection of two simple queries, $\den{r_1(e_1) \sqcap r_2(e_2)} = \{e_i \mid (e_i, r_1, e_1) \in \sK \land (e_i, r_2, e_2) \in \sK \}$.

\subsection{Simple Questions} \label{subsec:simple_quest}

Fig.~\ref{fig:figure2} provides an overview of our procedure for creating \emph{simple question} examples: (i) We manually define query templates, 
(ii) populate query templates using $\sK$ to create queries with a sufficiently large number of answers in $\sK$, (iii) automatically identify evidence passages for the answers and filter out noisy examples, (iv) map query templates to question templates to obtain pseudo-language questions, and (v) validate answers and paraphrase pseudo-language questions through crowdsourcing. Next, we describe each of these steps in detail.

\paragraph{Generating query templates}

We manually select a set of 135 relations $\bar{\sR} \subset \sR$, which will be used in our query templates. We select frequent relations from Wikidata for which denotations contain many entities (e.g., \texttt{ReceivedAward}).
The list of relations is in App.~\ref{sec:appendix_props}. For each relation, we manually write a template to map queries to pseudo-language questions. For example, the template for \texttt{ReceivedAward} is \emph{``Who received the award X?''}

Some relations are underspecified -- for example, \texttt{LocatedIn} can describe the location of buildings, geographical features, and cities. When generating synthetic questions, this leads to vague questions such as \emph{``What is located in Paris?''}. To address this, we manually split these to \emph{typed relations} that specify the semantic type of their answers/denotations. This is done using the type hierarchy given in Wikidata and given the type $t$ of answer entities. We denote typed relations by $r_t$, and the denotation of $r_t(e)$ comprises all entities of type $t$ returned by $r(e)$.
For example, the entity \texttt{The Louvre} has type \texttt{cultural organization}, and we can 
map the relevant query template to the pseudo-language question \emph{``Which cultural organization is located in Paris?''}. 

\paragraph{Simple query generation} 
We instantiate all possible simple queries using all $r \in \bar{\sR}$ and entities $e$ in Wikidata. For a relation $r$ (or $r_t$), we keep the query $r(e)$ iff $|r(e)| \geq 5$. We denote this set of instantiated simple queries by $\mathcal{S}$, which contains 1,431,268 simple queries.

\paragraph{Finding evidence sentences}
For an ODQA benchmark, we must verify that every answer is found in our target corpus. We do this by identifying candidate evidence sentences from Wikipedia, and verifying they entail the answer, using a Natural Language Inference (NLI) model.

Specifically, every simple query-answer pair can be viewed as a triple $(e_1, r, e_2)$. We use a ``distant supervision'' approach \cite{mintz-etal-2009-distant}, similar to KELM \cite{kelm}, and define any sentence in the Wikipedia page of entity $e_1$ that contains the entity $e_2$, or one of its Wikidata aliases, as a candidate evidence sentence (and vice versa in the page of $e_2$).
E.g., in Fig.~\ref{fig:figure2}, the evidence for (\texttt{BarackObama}, \texttt{ReceivedAward}, \texttt{NobelPeacePrize}) appears on the page \texttt{Barack Obama}, where \textit{`Nobel Peace Prize'} appears. 

Aligning Wikipedia sentences to Wikidata can lead to false positives. E.g., for the triple (\texttt{TheGoonies}, \texttt{HasScreenwriter}, \texttt{StevenSpielberg}), most mentions of Spielberg in the page
\texttt{TheGoonies} are not as a screenwriter. To account for this, we use an off-the-shelf NLI model.\footnote{huggingface.co/ynie/roberta-large-snli\_mnli\_fever\_anli\_R1\_R2\_R3-nli} For every answer, we consider each candidate evidence sentence along with its two preceding sentences, and check whether they entail the hypothesis phrase describing the triple $(e_1, r, e_2)$. We use templates to phrase triples as short declarative sentences (\emph{``The Goonies has Steven Spielberg as screenwriter'')}. An answer is \emph{validated} if there is an evidence sentence that entails the triple.
Manual analysis shows
this process eliminates 70\% of false positives, while removing only 7.5\% of the correct alignments.

\paragraph{Query filtering} 
After finding evidence sentences, we only keep queries that at least 80\% of their answers were validated and their number of validated answers is between 5 and 200. The resulting set contains 60,792 simple queries, where each query has a set of validated answers, $\sA$, and of passages $\sP$ that contain the identified evidence sentences.\footnote{We keep a single evidence passage for every triple.}

\subsection{Complex Questions}
\label{subsec:complex_questions}

To increase diversity, we expand simple queries to composition and intersection queries, for which answers require reading two passages.

\paragraph{Intersection}
Intersection queries are generated by finding two simple queries such that the size of the intersection of their denotations is at least 5. To avoid improbable questions such as \emph{``Which competition was won by Manchester City and had Manchester City as a participant?''}, we add a constraint that the denotation of one of the simple queries cannot be a subset of the other. 
Formally, the set of intersection queries are all queries $r_1(e_1)\sqcap  r_2(e_2)$ such that $|\den{r_2(e_2) \sqcap  r_1(e_1)}| \geq 5$, $\den{r_1(e_1)} \nsubseteq \den{r_2(e_2)}$, and $\den{r_2(e_2)} \nsubseteq \den{r_1(e_1)}$.
      
Pseudo-language questions are generated by heuristically combining the two simple questions, for example \textit{``Which television program had Chris Carter as screenwriter and had Frank Spotnitz as screenwriter?''}. There is no need to perform answer validation since all of the underlying intersecting answers were already validated.

\paragraph{Composition} 
To create composition queries, we manually handpick a set of 423 relations $\sR_{\text{comp}} \subset \sR$ (list in our codebase), in a process similar to simple queries. Then, we generate all the possible composition queries $r_2(r_1(e))$ such that $r_1(e) \in \mathcal{S}$, $r_2 \in \sR_{\text{comp}}$, and $|\den{r_2(r_1(e))}| \geq 5$. An example composition query is \emph{``What is the height of buildings located in Dubai?''}.

Unlike intersection queries, in composition queries we need to validate that our new triples 
$(e_i, r_2, e_j)$, where $e_j \in \den{r_1(e)}$, are indeed supported by Wikipedia sentences. We use the same procedure to find evidence sentences for triples $(e_i, r_2, e_j)$, and consider an answer $e_i$ as 
\emph{validated} if both $(e_i, r_2, e_j)$ and $(e_j, r_1, e)$ can be aligned to Wikipedia. We keep all complex queries where 80\% of the answers are validated. 
Finally, we manually define templates for relations in $\sR_{\text{comp}}$ to generate pseudo-language questions.

\subsection{Questions from Wikipedia Tables}
\label{subsec:wikipedia_tables}
To further diversify \ourbenchmark{}, we create an analogous pipeline for generating simple and composition questions from Wikipedia tables, with more open-ended relations compared to Wikidata. We briefly describe this pipeline.

We look at all Wikipedia tables with title \emph{``List of X''} that have at least 5 rows, in total, 1,897 tables. We find the ``key'' column, $c_\text{key}$ in each table using the table classifier from \newcite{multimodal_qa}, which outputs the column of entities that the table describes. For example, in the table \emph{List of nuclear whistle blowers}, $c_\text{key}$ is \emph{`name'} and specifies the whistle-blower names.
This naturally creates simple questions of the form \emph{``Who or what is X?''}.

Simple questions are expanded to composition questions by looking at non-key columns, $c_\text{non-key}$ and asking what rows in the table have the value $v$ in column $c_\text{non-key}$. For example, what is the value in the column \emph{`Year'} for nuclear whistle-blowers.

Questions from Wikipedia are validated using a procedure similar to Wikidata. For each answer entity $e$, we validate that the Wikipedia page for $e$ contains the relevant words that are part of the name of the table as well as the value (for composition questions), and only keep questions where 80\% of the table rows are validated and the number of validated answers is at least 5. Overall, we generate 170 simple questions and 6,036 composition questions using this process.

\subsection{Data Split}
\label{sec:data_split}
\ourbenchmark{} contains a training set, whose goal is to teach the model to handle multi-answer questions. However, we do not want the model to memorize how particular Wikidata relations map to text patterns.
Consequently, we perform a \emph{relation split}, randomly splitting the set $\bar{\sR}$ into two equally-sized sets $\bar{\sR}_{\text{train}}$ and $\bar{\sR}_{\text{test}}$. Simple queries are assigned to the train/test set based on their relation, composition queries $r_2(r_1(e))$ are assigned to the test set iff either $r_1$ or $r_2$ are in $\bar{\sR}_{\text{test}}$, and intersection queries $r_1(e_1) \sqcap r_2(e_2)$ are placed in the test set iff both $r_1$ and $r_2$ are in $\bar{\sR}_{\text{test}}$.

We now create the train/development/test split (Tab.~\ref{tab:global_statistics}). The main bottleneck in our example generation pipeline is validation of the test set through crowdsourcing (\S\ref{subsec:mturk}), 
since each question requires validating all of the answers.
Thus, we pre-determine the test set to contain 1,000 simple questions (830 from Wikidata, 170 from Wikipedia tables) and 1,000 complex questions (400 Wikidata composition questions, 400 Wikidata intersection questions, 200 Wikipedia tables composition questions). For simple Wikidata questions, we sample 830 questions such that the distribution over relations from $\bar{\sR}_{\text{test}}$ is roughly uniform. All Wikipedia tables simple questions are placed in the test set, and for complex questions we randomly sample the pre-determined number from the set of generated questions. Last, the test set is randomly split in half to a development set and test set. We also sub-sample training set examples, such that each relation appears in at most 1,000 examples.

\subsection{Crowdsourcing}
\label{subsec:mturk}

\paragraph{Correctness validation}
For every question and answer, we present a crowdsourcing worker with the question, the answer, and links to the Wikipedia page (or pages for complex questions) with the evidence passage. We ask the worker to check if the question can be answered from the given pages, using the text only (no infoboxes or tables).

Since the vast majority of examples are correct, we test worker performance by injecting wrong answers in 10\% of the cases and reject workers that fail to identify wrong answers. Moreover, we manually verify 5\% of examples marked as \emph{correct} and all examples marked as \emph{incorrect}, and again reject low-performing workers. 
Overall, 24 annotators validated 30,259 answers for an average pay of 12.5\$ per hour. We find that our process for generating examples is accurate, with 96.6\% of the answers validated. Non-validated questions were replaced until 2,000 questions were validated. A question is defined non-validated if its number of distinct answers goes below 5. Snapshots from the presented tasks are in App. \ref{sec:mturk_val}.

\paragraph{Paraphrasing}
Since our questions are in pseudo-language,
we follow past work \cite{wang2015overnight} and ask workers to re-phrase 3,000 questions in the training set and the entire development/test set. 
We restrict this task to US or UK workers who pass a qualification test. We randomly verified half of the paraphrases for each worker for quality assurance.
\section{Dataset Analysis}
\label{sec:data_analysis}

\begin{table*}[t]
  \centering
  \scriptsize
  \renewcommand{\arraystretch}{1.2}
  \begin{tabular}{c c c| c c c c  c}
  \toprule
    & & \textbf{Total} &\textbf{Simp. WD} & \textbf{Simp. WP} & \textbf{Inter. WD}  & \textbf{Comp. WD} & \textbf{Comp. WP} \\
    \midrule
   \multirow{2}{*}{\textbf{\# Questions}} & train & 61,911 & 28,574 & - & 2,301 & 25,200 & 5,836 \\
    & dev +  test & 2,000 & 830 & 170 & 400 & 400 & 200\\ \hline
    \multirow{2}{*}{\textbf{Mean \# Answers}} & train & 13.25 & 16.65 & - & 9.19 & 9.74 & 13.35 \\
    & dev +  test & 13.23 & 15.69 & 23.84 & 8.94 & 8.77 & 11.51\\ \hline
    \multirow{2}{*}{\textbf{Median \# Answers}} & train & 8.0 & 9.0 & - & 7.0 & 7.0 & 8.0 \\
    & dev +  test & 7.0 & 7.5 & 17.0 & 7.0 & 6.0 & 7.0 \\\hline
    \multirow{2}{*}{\textbf{Mean Question len.}} & train & 12.69 & 8.78 & - & 16.69 & 15.18 & 19.47 \\
    & dev +  test & 9.51 & 7.91 & 8.61 & 11.65 & 10.35 & 10.99 \\ 
     \bottomrule
  \end{tabular}
  \caption{\ourbenchmark{} questions breakdown by their type (\textbf{Simp}le, \textbf{Inter}section or \textbf{Comp}osition questions) and underlying data source (\textbf{WD} for Wikidata, \textbf{WP} for Wikipedia tables).}
  \vspace{-6mm}
  \label{tab:global_statistics}
\end{table*}

\ourbenchmark{} contains 61,911 training examples, 1,000 development examples and 1,000 test examples. Tab.~\ref{tab:examples} provides example questions of each question type and data sources.
We describe key statistics in Tab.~\ref{tab:global_statistics}. Test examples in \ourbenchmark{} have 13.23 answers on average and a median of 7 answers. For comparison, the number of answers per question is substantially higher than in AmbigQA \cite{ambig_qa}, where the median is 2. On average, simple questions have more answers than complex ones while being shorter in length. We note that since test and development questions were manually re-phrased by annotators they are generally shorter than the training questions.

\begin{figure}[t]
    \centering
    \subfloat[\centering \# answers]{{\includegraphics[width=3cm]{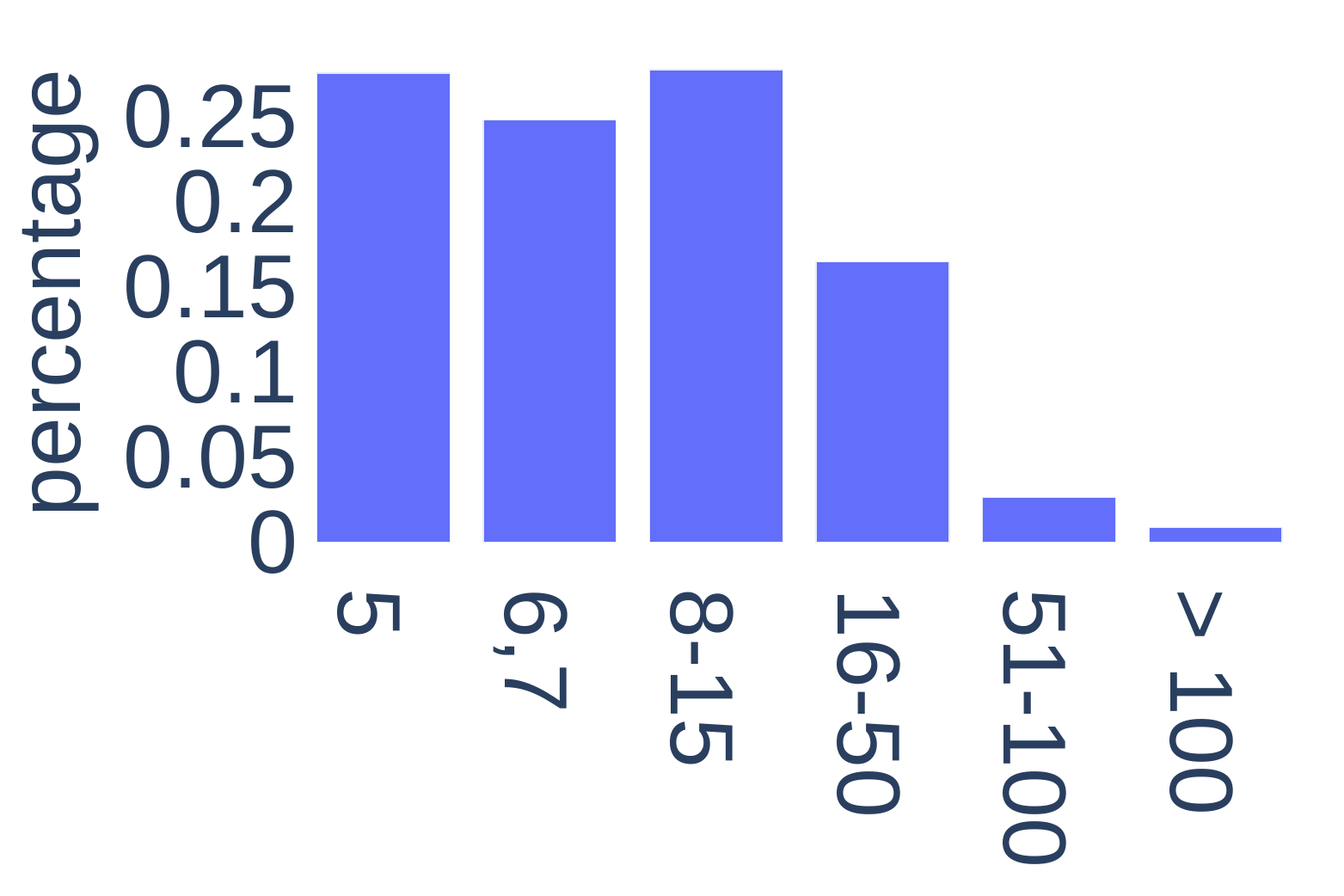}}}%
    \qquad
    \subfloat[\centering \# added answers]{{\includegraphics[width=3cm]{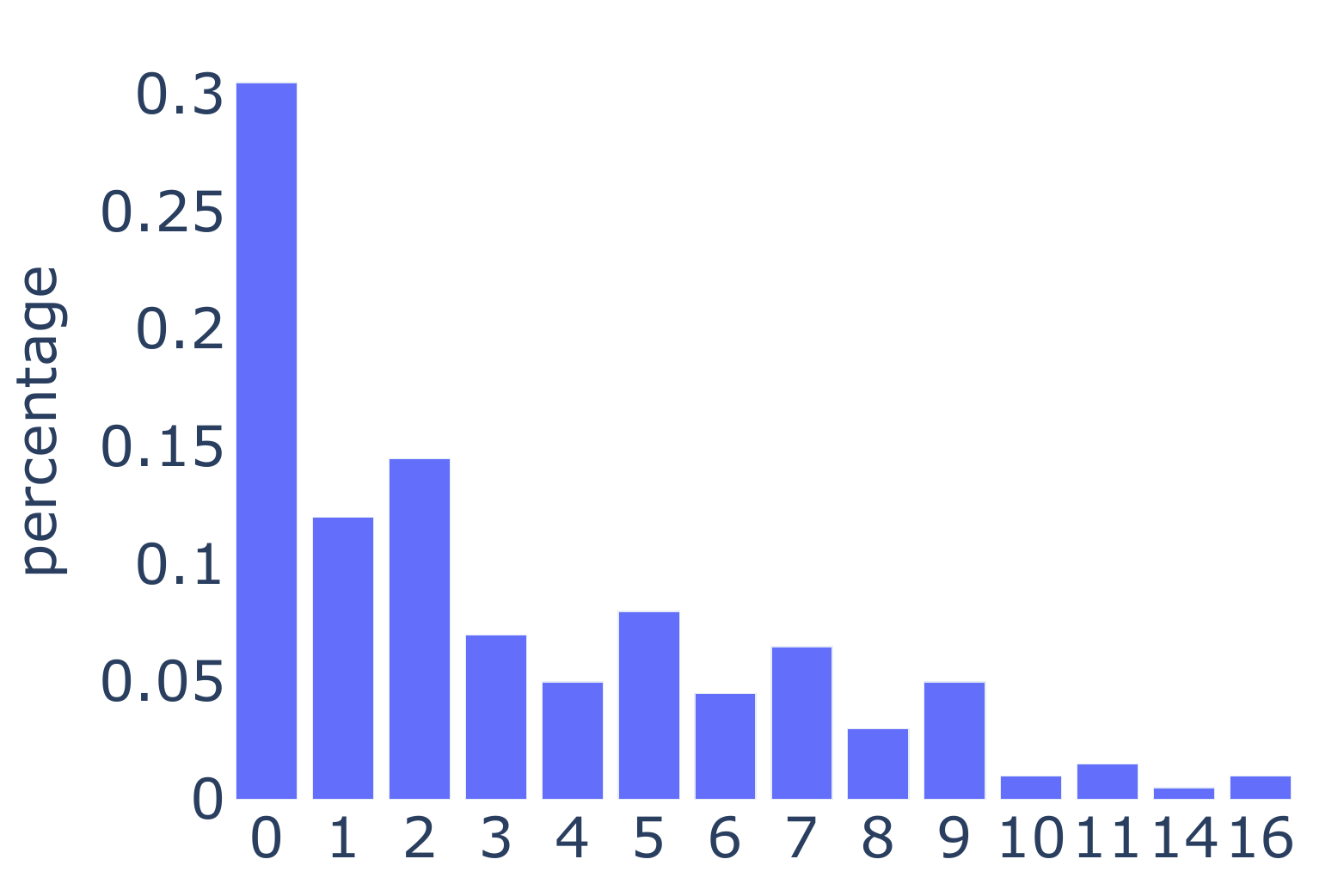} }}
    \caption{Left: Distribution of the number of answers per example. Right: Proportion of questions per number of added answers in  \emph{ExtendedSet}.}
    \vspace{-6mm}
    \label{fig:num_ans_proofs_dist}
\end{figure}

Figure \ref{fig:num_ans_proofs_dist}a presents a binned distribution of the number of answers per question in the development and test sets.
Roughly half of the questions have 8 or more answers, with 20\% having more than 15 answers and 3.5\% with over 50 answers.

\paragraph{Extended set}
\label{subsec:extended_set}

As discussed in \S\ref{subsec:mturk}, we manually validate each answer in \ourbenchmark{} is supported by sentences from Wikipedia. However, Wikipedia might contain additional correct answers.
To alleviate this issue, we manually annotate additional gold answers for a subset of test questions, and name it the \emph{ExtendedSet}.
We randomly sampled 200 questions from the test set
and had an author manually annotate as many additional answers as possible in 12 minutes, per question. This process is not guaranteed to be complete, as it would require manually reviewing all of Wikipedia. Moreover, questions with hundreds of gold answers (\emph{``Who worked for Burton F. C?''}) would incur hours of annotation, which is too expensive. This is similar to work in open information extraction \cite{vo2017open}, where creating the full gold set of triples is not feasible. 
Fig.~\ref{fig:num_ans_proofs_dist} plots the number of added answers per question on the extended set. In 30\% of the questions, we did not add any answer, and the median/average/maximum number of added answers are 2/3.13/16 respectively. Evaluation on the test set and the extended set in \S\ref{subsec:results} shows that model precision on the extended set is somewhat higher, but does not alter model ranking, illustrating the reliability our test set.
\section{Experimental Evaluation}
\label{sec:experiments}

\subsection{Models}
\label{subsec:models}

\paragraph{Retriever}
For retrieval, we experiment with both sparse and dense retrieval models on Wikipedia. As discussed in \S\ref{sec:dataset_concstrution}, we chunk Wikipedia into passages of consecutive sentences, using NLTK's sentence tokenizer, where each passage is 100 words on average. For all retrievers, we evaluate retrieval accuracy of the top-200 passages returned per question.

We use BM25 \cite{robertson2009bm25} as a strong sparse retrieval model. BM25 scores question-passage pairs based on their lexical similarity. It has been shown that BM25 is notoriously hard to beat using unsupervised retrieval methods \cite{izacard2021towards,ram2022spider}, and achieves comparable performance to that of supervised methods \cite{thakur2021beir}.
As our dense retriever we finetune on \ourbenchmark{} a DPR model \cite{dpr_article} trained on NQ. 
We finetune DPR in the typical contrastive manner (in-batch training), with one positive and one negative passage per question. Positives are sampled from the evidence passages, and negatives are sampled from the top-10 highest scoring passages, according to BM25, which do not contain the answer.

%DPR is trained in a contrastive manner, with positive examples and negative examples. We use contexts returned by BM25 that do not contain any of the answers as our negative examples, and use the proofs our pipeline provides for each answer as our positive examples. %DPR encodes questions and passages and uses their vector embeddings to compute their similarity.

\paragraph{Reader}

We experiment with two readers -- a Passage-Independent Generator (PIG), which reads each passage independently (a-là RAG \cite{RAG}), and a Fusion-in-Decoder (FiD) model \cite{fid_article}, which reads multiple passages simultaneously.

PIG is an encoder-decoder model that takes each of the retrieved passages as input and decodes a single answer or outputs \emph{``Not Relevant''} to indicate there is no answer. The final output is the union of all decoded answers across retrieved passages. We initialize PIG with T5-large \cite{t5-paper} and train with standard maximum likelihood.  
We use evidence passages as positive examples and the top scoring retrieved passage that is not an evidence passage and does not contain an answer (or its aliases) as a negative example.
%BM25 or DPR passages that are not part of the evidences as negative context to train PIG to generate \emph{``Not Relevant''}. The final answer is the union of all answers predicted accross passages.

FiD encodes each of the retrieved passages along with the input question. Its decoder then attends to the encoded representation and outputs a list of answers. We initialize FiD using a pretrained T5-Large model \cite{t5-paper} and train with standard maximum likelihood.

%\edit{For a reader model over multiple passages, we use a Fusion-in-Decoder (FiD) model \cite{fid_article}. 
%FiD encodes each of the retrieved passages along %with the input question. Its decoder then attends to the encoded representation and outputs a list of answers. We initialize FiD using a pretrained T5-Large model \cite{t5-paper} and train with standard maximum likelihood. }{We use a Fusion-in-Decoder (FiD) model \cite{fid_article} and a Passage-Independent Generator (PIG), a-là RAG \cite{RAG} as reader models over multiple passage. 
%FiD encodes each of the retrieved passages along with the input question. Its decoder then attends to the encoded representation and outputs a list of answers. 
%PIG, on the contrary, encodes and decodes each passage independently.}

%\edit{}{We train PIG to output for each passage either \emph{``Not Relevant''} for no answer, or \emph{``Answer: X''} for some \emph{X}. We use the evidence passages as positive contexts to train PIG to emit \emph{``Answer: X''} and we use the top scoring BM25 or DPR passages that are not part of the evidences as negative context to train PIG to generate \emph{``Not Relevant''}. The final answer is the union of all answers predicted accross passages.}
FiD is computationally expensive, as its decoder attends to a large number of encoded tokens and the generated output is long. Thus, we can only fit the top-50 passages returned by the retriever on a single A100 GPU.

%During training we apply teacher forcing \cite{williams1989learning}, i.e., the model is first given the gold passages followed by the top scoring passages according to the retriever. If $|\sP| > 50$, the model sees a random sample of size 50 from the set of gold passages.

\paragraph{Closed-book question answering}
We also experiment with a closed-book setting, where the QA model generates answers from knowledge encoded in its parameters without any evidence passages. We initialize our closed-book QA model with T5-SSM with 3B parameters \cite{roberts-etal-2020-much}, and train it with standard maximum likelihood -- the question is provided as input, and the model is trained to generate the gold set of answers. 

\paragraph{Zero-shot}
We test the zero-shot ability of Open AI's \emph{text-davinci-003}, from the Instruct-GPT family \cite{gpt3instruct}. We use GPT-3 in: (a) closed-book QA setup; (b) as a multi-passage reader. In the closed-book setup, the model receives only the question and is asked to provide a list of answers. In the reader setup, the model gets the question and the 15 highest-ranking passages from BM25 (the maximal number that fits in the context) and is asked to output a list of answers.

%This performance seems better than that of fully supervised question decomposition models \cite{quest_decomp_mor, perez-unsupervised-quest-decomp}. While further improvements to the QD model may be possible we leave them as future work.
\subsection{Experimental Setup}
\label{subsec:setup}

We created \ourbenchmark{} as a benchmark to be evaluated alongside additional ODQA benchmarks, such as NQ. Since it is semi-automatically generated, one can develop models tailored for \ourbenchmark{}. However, our goal is to have a single model that performs well across a wide variety of question types. Thus, we train and test models in a multi-task setup, on both NQ and \ourbenchmark{}, in addition to a \ourbenchmark{} only setting. We also train our models on NQ only and evaluate them on \ourbenchmark{}, to verify \ourbenchmark{}'s training set indeed improves answering questions with many answers.

Our main metrics are recall, precision, and F$_1$. Specifically, for test example $(q, \sP, \sA)$, and a predicted set of answers $\sA_\text{pred}$, recall, precision, and F$_1$ are standardly computed by comparing $\sA$ and $\sA_\text{pred}$, allowing for aliases (i.e., a gold answer is covered if it or one of its aliases are in $\sA_\text{pred}$). The model scores are  averaged across examples.
To get a sense of the average accuracy across examples, we measure the fraction of examples with F$_1$ of at least 0.5 (\%F$_1$ $\geq$0.5) and the fraction with recall of at least 0.8 (\%Recall$\geq$0.8). For NQ, we report the standard exact match (EM) metric.

We evaluate the retriever with \textsc{Recall@K}, that is,
the fraction of answers that appear in the top-K retrieved passages, averaged across examples. This metric comes in two flavors: (a) Answer \textsc{Recall@K} (\textsc{ARecall@K}): for every gold answer whether it or one of its aliases appear in the top-K retrieved passages. It is a loose metric since an answer can appear in a passage that does not provide any evidence to support the answer; (b) Evidence \textsc{Recall@K} (\textsc{ERecall@K}): since we have evidence paragraphs for every answer, we consider for every gold answer the fraction of evidence passages in the top-K retrieved passages. This is a strict metric since an answer can sometimes be answered by passages other than the ones we identified.

\subsection{Results}
\label{subsec:results}

\begin{table}[t]
  \centering
  \scriptsize
  \renewcommand{\arraystretch}{1.2}
  \begin{tabular}{ l c c c c}
  \toprule
    & \multicolumn{2}{c}{\textbf{ARecall@K}} & \multicolumn{2}{c}{\textbf{ERecall@K}} \\
    & \textbf{\textsc{BM25}} & \textbf{\textsc{DPR}} & \textbf{\textsc{BM25}} & \textbf{\textsc{DPR}} \\
    \midrule
    K=10 & 24.6 & 21.9 & 11.1 & 11.1 \\
    K=25 & 37.4 & 31.5 & 28.4 & 16.2 \\
    K=50 & 46.6 & 39.6 & 38.7 & 20.8 \\
    K=100 & 54.6 & 47.1 & 47.6 & 25.5 \\
    K=200 & 61.0 & 55.2 &  55.6 & 30.2 \\ \bottomrule
  \end{tabular}
  \caption{Retriever test results.}
  \label{tab:retriever_baselines}
  \vspace{-6mm}
\end{table}

Tab.~\ref{tab:retriever_baselines} presents passage retrieval results on \ourbenchmark{} test. Scores for ARecall@200 for BM25 and DPR are 61.0\% and 55.2\%, respectively. As for ERecall@K, results are unsurprisingly lower. BM25 retrieves 55.6\% of the evidence passages with K=200, while DPR retrieves only 30.2\% of evidence passages.\footnote{While ERecall@K for DPR is substantially lower than BM25,
observe that Arecall@K is better correlated with QA metrics (Tab.~\ref{tab:reader_baselines}), as DPR retrieves non-evidence passages that still lead to the correct answer.}
Overall, DPR pretrained on NQ and finetuned on QAMPARI performs worse than BM25.
This is in line with \citet{entity-centric} who showed that, when tested on questions with \emph{rare entities}, DPR performs worse than BM25. We hypothesize that rare entities in \ourbenchmark{} questions may account for DPR's lower performance.

\begin{table}[t]
  \centering
  \scriptsize
  \renewcommand{\arraystretch}{1.1}
  \begin{tabular}{l @{\hspace{0.7\tabcolsep}} c @{\hspace{0.7\tabcolsep}} c @{\hspace{0.7\tabcolsep}} c @{\hspace{0.7\tabcolsep}} c @{\hspace{0.7\tabcolsep}} c @{\hspace{0.7\tabcolsep}} c}
  \toprule
    & & \textbf{Rec.} & \textbf{Prec.} & \textbf{F$_1$} & \textbf{\%Rec$\geq$.8} & \textbf{\%F$_1\geq$.5} \\
    \midrule
    \multirow{2}{*}{\textbf{FiD-BM25}} & QO & 25.1 & 36.8 & 28.3 & 6.8 & 24.2 \\
    & MT & 26.9 & 37.7 & 29.7 & 7.4 & 25.6 \\  
    \multirow{2}{*}{\textbf{FiD-DPR}} & QO & 7.8 & 39.1 & 12.5 & 0 & 3.6 \\
    & MT & 7.8 & 41.3 & 12.5 & 0 & 2.6 \\ 
    \hline
    \multirow{3}{*}{\textbf{PIG-BM25}} & NQO & 34.6 & 19.3 & 20.8 & 18.5 & 11.9 \\
    & QO & 43.1 & 30.7 & 31.0 & 26.7 & 26 \\
    & MT & \textbf{47.9} & 28.2 & 30.5 & \textbf{31.2} & 22.3 \\
    \multirow{3}{*}{\textbf{PIG-DPR}} & NQO & 9.0 & 13.7 & 8.4 & 0.5 & 2.6 \\
    & QO & 36.2 & 41.1 & \textbf{32.8} & 15.7 & 30.7 \\
    & MT & 34.1 & \textbf{44.8} & 32.4 & 15 & \textbf{31.3} \\ 
    \midrule
    \textbf{Closed book} & ZS & 12.9 & 17.4 & 13.8 & 1.9 & 9.5 \\
    \textbf{Reader} & ZS & 20.0 & 22.8 & 18.8 & 5.8 & 13.8 \\
    \midrule
    \textbf{Closed book}& QO & 1.7 & 7.3 & 2.6 & 0 & 0.3  \\
     \bottomrule
  \end{tabular}
  \caption{\ourbenchmark{} test results. \textbf{QO}: models trained on \ourbenchmark{} only; \textbf{NQO}: models trained on NQ only; \textbf{MT}: Multi-task training with NQ; \textbf{ZS}: Zero-shot setup. }
  \vspace{-6mm}
  \label{tab:reader_baselines}
\end{table}

Tab.~\ref{tab:reader_baselines} lists results on the test sets of \ourbenchmark{} and NQ. Overall, performance on \ourbenchmark{} is low. FiD-DPR and PIG-DPR are more precision-oriented with FiD-DPR achieving precision of 41.3 and PIG-DPR a precision of 44.8. PIG-BM25 is recall-oriented, achieving recall of 47.9. 
Overall, PIG variants perform best, with small differences between PIG-BM25 and PIG-DPR, and both are slightly higher than FiD-BM25. 

When training on both NQ and \ourbenchmark{} (MT), performance on NQ (47.2 with BM25 and 53.1 with DPR) is similar to that reported by \newcite{fid_article} (44.1 with BM25 and 51.4 with DPR). When training on NQ only, results on \ourbenchmark{} are significantly lower than when training also on \ourbenchmark{}, showing that training on \ourbenchmark{} improves performance on multi-answer questions, as expected.
The lower performance on \ourbenchmark{} compared to NQ, despite the fact that NQ's EM evaluation metric is much more strict than the metrics used for \ourbenchmark{}, illustrates the challenge in answering multi-answer questions.

PIG-DPR has much higher recall than FiD-DPR, showing that going over 200 passages independently (PIG) leads to higher recall than jointly reasoning over 50 passages (FiD). Moreover, the solid performance of PIG-DPR  indicates that QA performance is more correlated with ARecall@K than ERecall@K (Tab.~\ref{tab:retriever_baselines}).

%Results using DPR are significantly lower than those using BM25 \edit{}{on FiD but not for PIG}. 

%\edit{This corresponds to our findings in Tab.~\ref{tab:retriever_baselines} that DPR-NQ retriever performs significantly worse than BM25.}{It highlights the asymmetry between reasoning over multiple paragraphs, versus finding the answer within a single paragraph.}

Finetuned closed-book performance is low with an F$_1$ of 2.6 for \ourbenchmark{}, which we attribute to the relation-based train/test split (\S\ref{sec:data_split}). This guarantees that there is no overlap between train and test questions. \citet{question-overlap} have shown that mitigating such train-test overlap causes a drop in QA performance, with a drastic drop being observed in closed-book models.

\paragraph{Zero-shot results}
The performance of zero-shot models is lower than finetuned retrieve-and-read models, as expected. However, \emph{text-davinci-003}'s performance in the closed book setup is impressive and significantly better than finetuned T5-3B.
%\subsection{Analysis}

%\edit{In \S\ref{subsec:precision_analysis} we note that questions may include additional answers in Wikipedia that are absent from \ourbenchmark{}. We estimate this effect on model precision by manually annotating 30 questions from the development set with additional answers that we find in Wikipedia. Evaluating FiD-BM25 (MT) on this set reveals its true precision to be 65.8\%, compared to 49\% when using only the \ourbenchmark{} answers. %originally-predicted answers. 
%This demonstrates that precision should be used to rank models on \ourbenchmark{}, but not as an absolute measure of true precision.}
%{In \S\ref{sec:ext_reader} we report results for FiD and PIG on the \emph{ExtendedSet} with BM25. We first note that the precision is improved across all models whereas recall is improved for FiD but not for PIG. However the absolute precision changes are larger than the absolute recall changes (around 6 points vs. 2 points). The last interesting point is that order between models is preserved between the \emph{ExtendedSet} with annotations and without annotations. Therefore, all metrics can be used to rank between models but recall and F1 are closed to the actual recall and F1 (on all Wikipedia) than precision.}

\paragraph{\emph{ExtendedSet} results}
We report results for FiD and PIG on the \emph{ExtendedSet} (see \S\ref{sec:data_analysis}) in \S\ref{sec:ext_reader}.
As expected, considering additional correct answers improves the precision of all models. Since changes to recall are small, the overall F$_1$ is higher when considering manual annotations.  %\edit{}{We suggest focusing on recall and F1 since they are less subject to additional annotation changes.} 
Importantly, ranking across models does not change, and the absolute performance remains low, suggesting that our test set can be safely used for evaluation.

\paragraph{Oracle analysis}

To disentangle retrieval from answer extraction, we run PIG and FiD in an oracle setup, where we assume a perfect retriever and run our readers on the gold evidence passages only. Performance of both models greatly improves in this setup, with larger gains for PIG. This shows that developing better retrieval mechanisms for multi-answer questions can greatly benefit \ourbenchmark{}.
FiD's recall is still limited (47.5), illustrating the challenge of reading a large number of documents.
Full oracle results are in \S\ref{sec:dev_reader} (Tab.~\S\ref{tab:reader_dev_baselines}).

%We report full oracle results in the Appendix (\S\ref{tab:reader_dev_baselines}). Since PIG is recall-oriented whereas FiD is precision-oriented, PIG benefits more than FiD. PIG's performance in the Oracle setup is quite high, showing that the retriever is the main problem for PIG. FiD's performance is better than in the standard setup, but ts recall is still limited, illustrating that reading simultaneously a large number of documents is challenging.

%We first note that the precision is improved across all models whereas recall is improved for FiD but not for PIG. However the absolute precision changes are larger than the absolute recall changes (around 6 points vs. 2 points). The last interesting point is that order between models is preserved between the \emph{ExtendedSet} with annotations and without annotations. Therefore, all metrics can be used to rank between models but recall and F1 are closed to the actual recall and F1 (on all Wikipedia) than precision.

\section{Related work}
% TW: NEW SHORTER VERSION - 14/12

ODQA tasks have largely been dedicated to single-answer questions \cite{berant2013freebase, trivia_qa, natural_question_qa}. The same applies for most multi-hop ODQA tasks \cite{welbl-etal-2018-constructing,hotpotqa,trivedi-etal-2022-musique}. While they require 2-4 paragraphs, the answer is a single phrase.
Multi-answer questions were introduced in the TREC QA tracks \cite{trec_qa_2003, trec_qa_2004}. However, evaluation was on 50 questions.
%The AmbigQA dataset \cite{ambig_qa} introduced ambiguous questions that have multiple answers, depending on their interpretation. In contrast, questions in \ourbenchmark{} are unambiguous and contain significantly more answers.
%In WikiNLDB \cite{nrot_qa} questions require that reasoning over sets of facts. However, its retrieval is restricted to a corpus of 1,000 model-generated sentences. Contrastly, we use Wikipedia as our open-domain corpus, making our setup more realistic and challenging. 
\citet{TeaBreaC-2022} introduced artificially generated multi-answer questions, but only for reading comprehension rather than ODQA.
Concurrent to \ourbenchmark{}, \newcite {RoMQa} introduced RoMQA, a benchmark containing multi-answer questions generated using Wikidata. While their setup is closest to ours, they evaluate on a subset of Wikipedia that is aligned to a subset of Wikidata. %Last, their extensive user study finds \ourbenchmark{} questions to be more natural than in RoMQA.

%\paragraph{Models}

Retrieve-and-read models are the prevailing approach in ODQA \cite{chen_2017_odqa,yang2019end,lee-etal-2019-latent,sachan2021emdr}. 
%When the number of evidence passages is large, such models must fetch all relevant passages in order to generate the answer. 
Closed-book QA is an alternative approach, \cite{roberts-etal-2020-much,tay2022transformer} but requires using high-capacity models. %A less explored approach, potentially suitable for large answer sets, is that of virtual knowledge-bases (KBs), which encode a corpus into a differentiable KB that is amenable for retrieval and logical operations \cite{reasoning_over_vkb, differentiable_vkb}.

\section{Conclusions}

We release \ourbenchmark{}, a dataset targeting ODQA models ability to answer multi-answer questions, and show that it is challenging for current state-of-the-art models.
% Multi-answer questions are an integral part of ODQA that has thus far been neglected. 
\ourbenchmark{} will aid  develop models that answer a wide range of question types, including single- and multi-answer questions.

\section*{Limitations}

A key limitation of \ourbenchmark{} is that the gold set of answers is incomplete. Thus, predicted answers might be correct but missing from the gold answer set. The \emph{ExtendedSet} addresses this problem partially, allowing a more accurate model ranking, but even in this set all the correct answers are not part of the gold set.
A second limitation is that our data generation process is mostly automatic and is thus amenable to reverse-engineering. Hence, we recommend evaluating models on \ourbenchmark{} along with additional benchmarks created with a different generation process.
Last, our data generation process can only generate answers based on relations from Wikidata and relations that are in Wikipedia tables, and thus its scope does not generalize to arbitrary relations.

%\section*{Acknowledgements}
%We want to thank Omer Bigi Amouyal, Levana Amouyal and Joseph McCrum for their help with the annotation verification process. We also want to thank Ori Ram for his helpful comments. 
%This research was supported in part by The Yandex Initiative for Machine Learning, and The European Research Council (ERC) under the European Union Horizons 2020 research and innovation programme (grant ERC DELPHI 802800). 

\bibliography{custom_rebibed}
\bibliographystyle{acl_natbib}
\appendix

\section{Simple Relations}
\label{sec:appendix_props}
 
\begin{table*}[t]
  \centering
  \tiny
  \renewcommand{\arraystretch}{1.1}
  \begin{tabular}{l l l l l}
    is a & has author & located in & language & occupation \\ 
    sex or gender & country of citizenship & part of & place of birth & located in \\ 
    educated at & language spoken, written or signed & has part & played the sport & employer \\ 
    genre & position held & cast member & country of origin & award received \\ 
    place of death & made from material & creator & has participant & depicts \\ 
    maintained by & operator & performer & member of political party & owned by \\ 
    religion & headquarter location & participant & member of & position played \\ 
    original language & competition class & publisher & role & record label \\ 
    work location & director & doctoral advisor & residence & native language \\ 
    place of publication & medical condition & winner & field of work & form or work \\ 
    conflict & place of burial & instrument & composer & league \\ 
    screenwriter & distribution format & producer & sponsor & ethnicity \\ 
    voice actor & distributed by & participating team & academic degree & manufacturer \\ 
    architectural style & fabrication method & present in work & production company & cause of death \\ 
    military branch & manner of death & industry & director of photography & narrative location \\ 
    original broadcaster & organizer & student of & location of creation & located in or next to body of water \\
    architect & archives at & nominated for & country of registry & allegiance \\ 
    movement & voice actor & noble title & based on & dedicated to \\ 
    legislated by & location of formation & developer & contributor to creative work or subject & lyrics written by \\ 
    located in protected area & tracklist & editor & presenter & religious order \\ 
    from narrative universe & location of discovery & media franchise & commissioned by & political ideology \\ 
    commemorates & port of registry & influenced by & indigenous to & operating area \\ 
    translator & brand & interested in & designed by & illustrator \\ 
    vessel class & costume designer & drafted by & coach of sports team & convicted of \\ 
    scenographer & culture & significant place & executive producer & represented by \\ 
    broadcast by & investor & cover art by & home port & collection creator \\ 
    armament & inspired by & first appearance & choreographer & animator \\ 
    source of energy & musical conductor & adapted by & sound designer & has written for \\ 
    academic major & ratified by & business model & worshipped by & narrator \\ 
    partnership with & colorist & art director & has work in the collection & military rank \\  
  \end{tabular}
  \caption{Simple relations}
  \label{tab:properties_table}
\end{table*}

 In Tab. \ref{tab:properties_table}. we gathered all the 135 relations we used to create our simple questions. The 423 relations used to create our composition questions can be found in our code base.
\section{Composition template}
\label{sec:composition_template}

Composition questions overall template is: \textbf{What is the $<$comp\_property$>$ of $<$subtype$>$ who/which $<$base\_property$>$?}. All the templates are in our code base.
\section{Crowdsourcing Validation}
\label{sec:mturk_val}

\begin{figure*}[t]
    \centering
    \subfloat[\centering Instructions]{{\includegraphics[width=10cm]{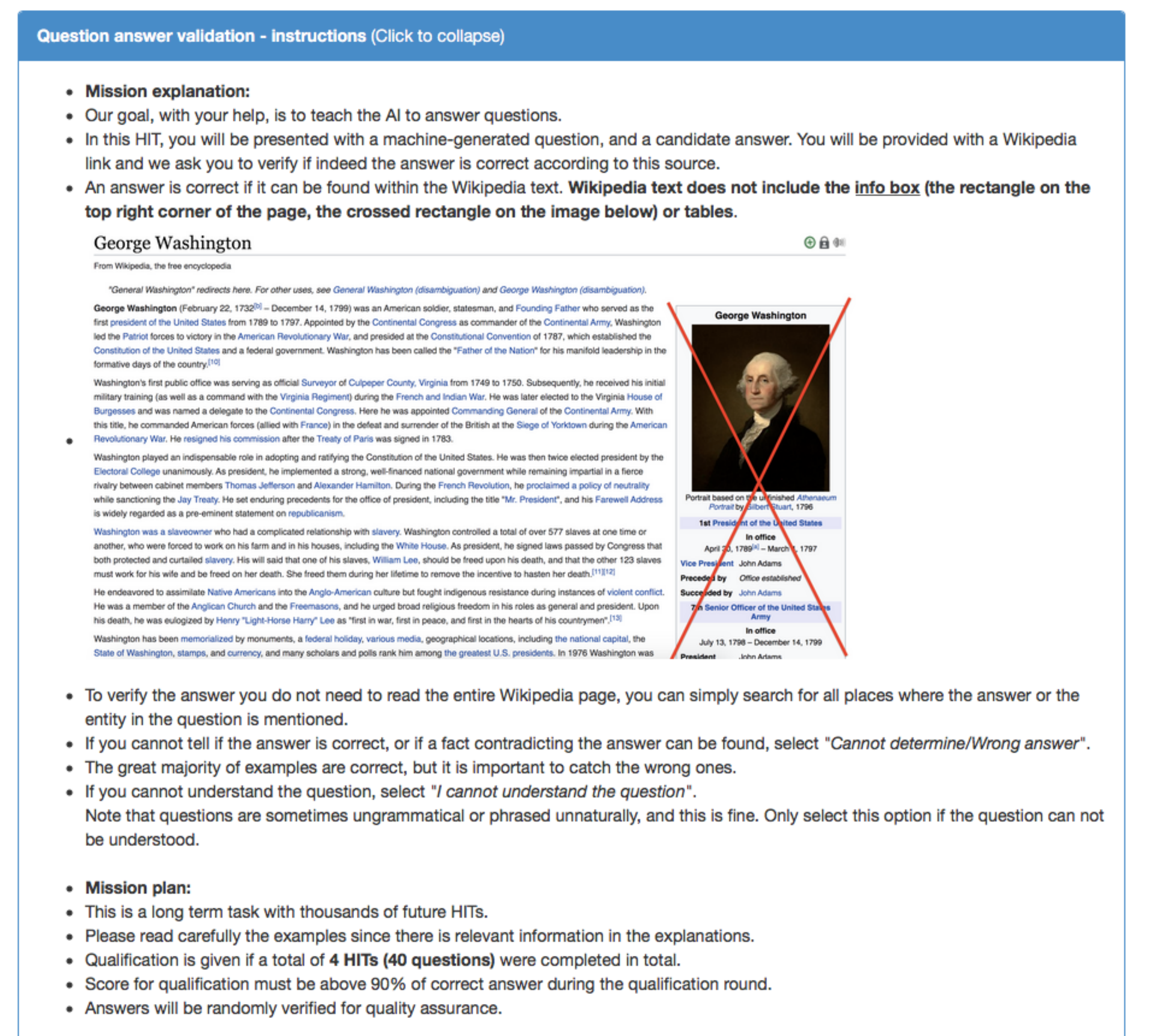}}}%
    \qquad
    \subfloat[\centering Task]{{\includegraphics[width=10cm]{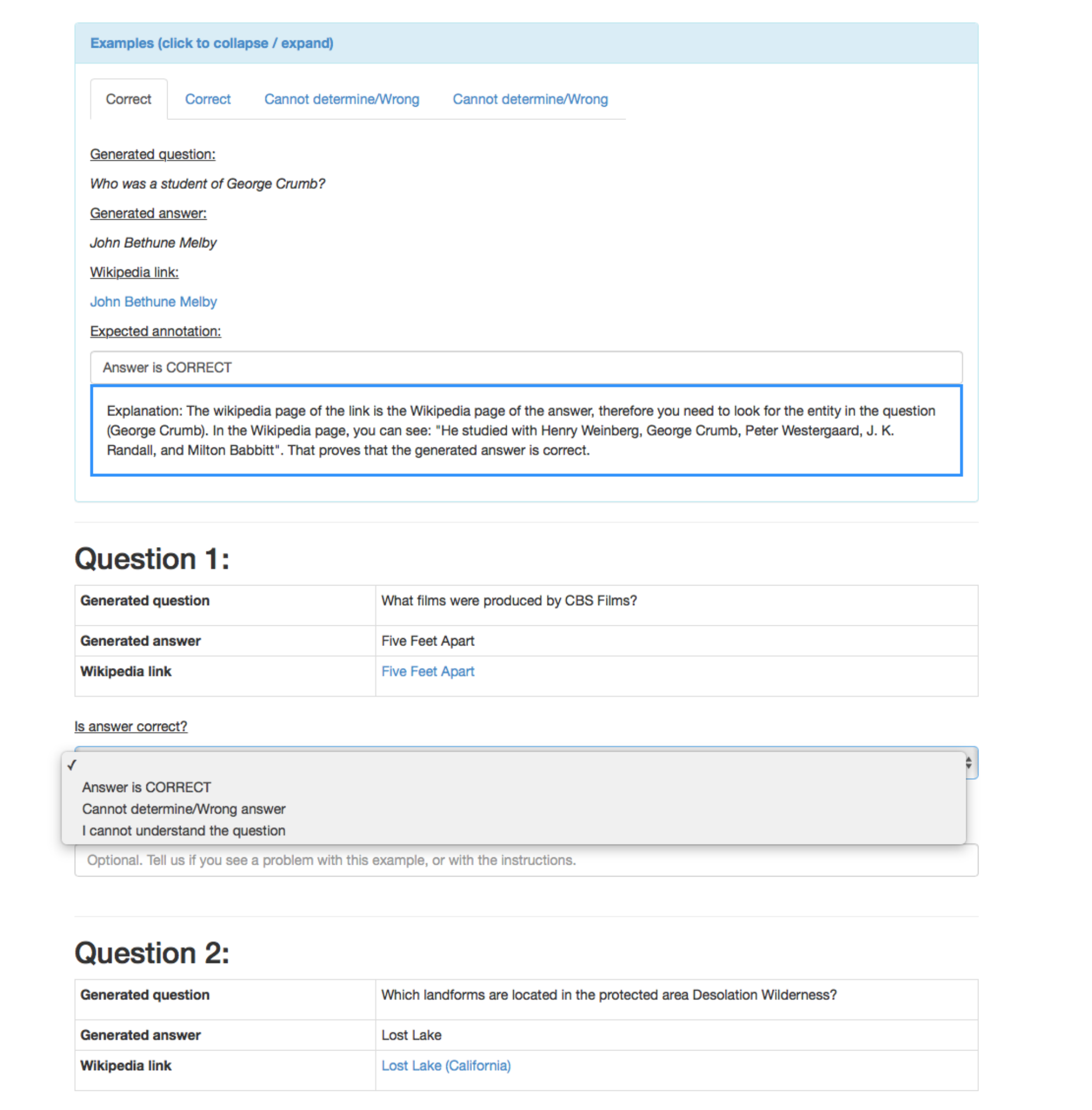} }}
    \caption{Screenshots from crowdsourcing task.}
    \label{fig:mturk_screenshot}
\end{figure*}

Fig.~\ref{fig:mturk_screenshot} shows two screenshots of the task crowdsourcing workers performed.

\section{Experimental setup details}
\label{sec:experiments_details}

For both readers (FiD and PIG), we used T5-large which has 770 million parameters. We used an A100 to train both of them, FiD with a batch size of 8 and PIG with a batch size of 32 for a single GPU. We trained each of them for around 48 hours on two GPUs. \\
For FiD, we concatenated the answers using $\#$ as a separator. At evaluation time, there is no importance to the order of the answers. \\
For both PIG and FiD, all aliases of a given
gold entity provided by Wikidata are used as additional correct answers. When verifying whether our model predicted an answer A, we verify
whether it predicted A or any of its aliases.
We performed an hyper parameter search around the learning rate, the number of training steps, the ratio of positive to negative (for PIG) and the number of times an NQ example will appear in each epoch (for multi task). Tab. \ref{tab:reader_hyper_params} presents the parameters of the reported results. \\
We report the results of a single run with seed 0.

\begin{table*}[t]
  \centering
  \footnotesize
  \renewcommand{\arraystretch}{1.1}
  \begin{tabular}{l c c c c c}
  \toprule
    & & \textbf{Learning rate} & \textbf{\# steps} & \textbf{pos. to neg.} & \textbf{\# NQ examples} \\
    \midrule
    \multirow{2}{*}{\textbf{FiD-BM25}} & QO & 0.00005 & 90k & - & - \\
    & MT & 0.00005 & 190k & - & 2 \\  
    \multirow{2}{*}{\textbf{FiD-DPR}} & QO & 0.00005 & 85k & - & - \\
     & MT & 0.00005 & 190k & - & 2 \\
    \multirow{2}{*}{\textbf{PIG-BM25}} & QO & 0.000001 & 60k & 1 & - \\
    & MT & 0.000001 & 75k & 1 & 1 \\  
    \multirow{2}{*}{\textbf{PIG-DPR}} & QO & 0.000001 & 60k & 1 & - \\
    & MT & 0.000001 & 75k & 1 & 1 \\
    \midrule
    \textbf{Closed book}& QO & 0.0001 & 95k & - & -  \\
     \bottomrule
  \end{tabular}
  \caption{Hyper parameters used for reported results.}
  \label{tab:reader_hyper_params}
\end{table*}
\section{Question type analaysis}
\label{sec:question_type_analysis}

We break test performance of FiD-BM25 (MT) by question type (Tab.~\ref{tab:FiD_in_depth}).
Surprisingly, performance on simple
questions is lower than complex
questions, and intersection questions
seem easiest. Possible explanations are: (a) simple questions have more answers (see Tab.~\ref{tab:global_statistics}), which makes them harder, and (b) models can predict the answer given just one evidence passage, due to ``shortcuts'' \cite{chen-durrett-2019-understanding}, or parametric knowledge \cite{longpre2021entity}.

\begin{table}[t]
  \centering
  \scriptsize
  \renewcommand{\arraystretch}{1.1}
  \begin{tabular}{l c c c}
  \toprule
    & \textbf{Recall} & \textbf{Precision} & \textbf{F$_1$} \\
    \midrule
    Wikidata simple & 21.3 & 30.7 & 23.1  \\
    Wikidata intersection & 37.0 & 47.1 & 40.0 \\
    Wikidata composition & 18.6 & 32.4 & 22.2  \\
    Wikipedia simple & 9.1 & 20.6 & 11.5  \\
    Wikipedia composition & 31.2 & 37.4 & 32.7 \\
     \bottomrule
  \end{tabular}
  \caption{Question type analysis of FiD-BM25, trained in MT setup on \ourbenchmark{} development set.}
  \label{tab:FiD_in_depth}
\end{table}
\section{\emph{ExtendedSet} Results}
\label{sec:ext_reader}

In Tab.~\ref{tab:reader_ext_baselines} we present results analogous to those in Tab.~\ref{tab:reader_baselines} for the \emph{ExtendedSet} with BM25.
Precision improves by 5-6 points across models, while recall changes are smaller leading to an overall increase in F$_1$. Nevertheless changes are not dramatic and model ranking remains constant, suggesting the full test set can be safely used.

\begin{table*}[t]
  \centering
  \footnotesize
  \renewcommand{\arraystretch}{1.1}
  \begin{tabular}{l c c c c c c}
  \toprule
    & & \multicolumn{5}{c}{\textbf{\ourbenchmark{}}} \\
    & & \textbf{Recall} & \textbf{Precision} & \textbf{F$_1$} & \textbf{\%Recall$\geq$0.8} & \textbf{\%F$_1$ $\geq$0.5} \\
    \midrule
    \multirow{2}{*}{\textbf{FiD-BM25 \hspace{0.5cm} QO}} & w.o. annotations & 20.5 & 34.6 & 24.3 & 4.0 & 19.6 \\
    & w. annotations & 23.3 & 40.6 & 27.8 & 4.5 & 25.1  \\
    \multirow{2}{*}{\textbf{FiD-BM25 \hspace{0.5cm} MT}} & w.o. annotations & 22.8 & \textbf{37.0} & 26.8 & 4.5 & 20.6  \\
    & w. annotations & 25.7 & \textbf{42.9} & 30.6 & 5.0 & 24.6 \\  
    \multirow{2}{*}{\textbf{PIG-BM25 \hspace{0.5cm} QO}} & w.o. annotations & 45.1 & 28.9 & \textbf{30.7} & 27.5 & \textbf{23} \\
    & w. annotations & 42.7 & 33.6 & 32.8 & 24 & \textbf{29.5}  \\
    \multirow{2}{*}{\textbf{PIG-BM25 \hspace{0.5cm} MT}} & w.o. annotations & \textbf{49.3} & 27.9 & \textbf{30.7} & \textbf{31.5} & 20.5  \\
    & w. annotations & \textbf{47.1} & 33.1 & \textbf{33.2} & \textbf{27} & 26 \\
     \bottomrule
  \end{tabular}
  \caption{\ourbenchmark{} \emph{ExtendedSet} results with (w.) and without (w.o.) the additional manual annotations. The best results with and without annotations are bolded. \textbf{QO}: models trained on \ourbenchmark{} only; \textbf{MT}: Multi-task training with NQ.}
  \label{tab:reader_ext_baselines}
\end{table*}
\section{Development Set Results}
\label{sec:dev_reader}

In Tab.~\ref{tab:reader_dev_baselines} we present results analogous to those in Tab.~\ref{tab:reader_baselines} for the development set.

\begin{table*}[t]
  \centering
  \footnotesize
  \renewcommand{\arraystretch}{1.1}
  \begin{tabular}{l c c c c c c}
  \toprule
    & & \multicolumn{5}{c}{\textbf{\ourbenchmark{}}}\\
    & & \textbf{Recall} & \textbf{Precision} & \textbf{F$_1$} & \textbf{\%Recall$\geq$0.8} & \textbf{\%F$_1$ $\geq$0.5} \\
    \midrule
    \multirow{2}{*}{\textbf{FiD-BM25}} & QO & 23.3 & 35.6 & 26.3 & 5.9 & 22.7 \\
    & MT & 23.9 & 34.2 & 26.3 & 6.0 & 22.4 \\  
    \multirow{2}{*}{\textbf{FiD-DPR}} & QO & 6.5 & 35.2 & 10.1 & 0 & 3.7 \\
    & MT & 7.2 & 39.8 & 11.4 & 0.0 & 2.8 \\
    \multirow{2}{*}{\textbf{PIG-BM25}} & QO & 41.4 & 26.4 & 28.0 & 25.3 & 21.0 \\
    & MT & 43.7 & 26.9 & 28.9 & 26.6 & 22.0 \\  
    \multirow{2}{*}{\textbf{PIG-DPR}} & QO & 33.9 & 38.6 & 29.9 & 15.8 & 26.2 \\
    & MT & 31.7 & 42.2 & 29.6 & 14.3 & 26.3 \\
    \midrule
    \textbf{Closed book}& QO & 2.4 & 7.2 & 3.1 & 0.1 & 0.7  \\
        \midrule
    \textbf{FiD-Oracle} & MT & 47.5 & 62.7 & 51.2 & 18.4 & 56.1 \\
    \textbf{PIG-Oracle} & MT & 71.5 & 60.9 & 62.4 & 55.7 & 73.8 \\
     \bottomrule
  \end{tabular}
  \caption{\ourbenchmark{} development results. \textbf{QO}: models trained on \ourbenchmark{} only; \textbf{MT}: Multi-task training with NQ. }
  \label{tab:reader_dev_baselines}
\end{table*}
\end{document}